\documentclass[runningheads]{llncs}
\usepackage{esvect}
\usepackage{marvosym}
\usepackage[T1]{fontenc} 
\usepackage[utf8]{inputenc} 

\usepackage[hidelinks, colorlinks=true, linkcolor=blue, citecolor=blue, breaklinks=true]{hyperref}
\usepackage{graphicx} 
\usepackage{epstopdf} 
\usepackage{multirow} 
\usepackage{xcolor} 
\usepackage{colortbl} 
\usepackage{booktabs} 
\usepackage{bm} 
\usepackage{ulem} 
\usepackage{pifont} 
\usepackage{amsfonts, amssymb, amsmath} 
\usepackage{esvect} 
\usepackage{inconsolata}
\usepackage{algorithm}
\usepackage{algpseudocode}
\usepackage{float}
\usepackage{longtable}
\usepackage{makecell}
\usepackage{latexsym}
\usepackage{enumitem}
\usepackage{color}
\usepackage{titletoc}
\usepackage{wrapfig}  
\usepackage{lipsum} 
\usepackage{listings}
\usepackage{array}
\usepackage{caption}  
\usepackage{multicol}
\usepackage{bbm}
\usepackage{dsfont} 
\usepackage{subcaption} 
\usepackage{ragged2e} 

\definecolor{myGreen}{RGB}{34, 139, 34}
\definecolor{myRed}{HTML}{FF6347}

\titlerunning{MedEndoSAM: Efficient Segmentation for Endoscopic Images}

\title{Stain-Aware Augmentation and Hybrid Loss for Domain Generalization for Robust Atypical Mitosis Classification}
\author{Anonymous}

\authorrunning{Anonymous}

\begin{document}
%
%
\titlerunning{Robust Atypical Mitosis Classification with DenseNet121}

\title{Stain-Aware Augmentation and Hybrid Loss for Domain Generalization for Robust Atypical Mitosis Classification}

\author{Adinath Dukre \and
        Ankan Deria \and
        Yutong Xie \and
        Imran Razzak \thanks{Corresponding author: imran.razzak@mbzuai.ac.ae}
}

\authorrunning{Dukre et al.}

\institute{Mohamed Bin Zayed University of Artificial Intelligence, Abu Dhabi, UAE \\
\email{\{adinath.dukre, ankan.deria, yutong.xie, imran.razzak\}@mbzuai.ac.ae}}

\maketitle              

\begin{abstract}
Atypical mitotic figures are important biomarkers of tumor aggressiveness in histopathology, yet their reliable recognition remains challenging due to severe class imbalance and variability across imaging domains. We present a DenseNet-121–based framework tailored for atypical mitosis classification in the \textsc{MIDOG-25} (Track~2) setting. Our method integrates stain-aware augmentation via Macenko normalization, geometric and intensity transformations, and imbalance-aware learning with weighted sampling, cost-sensitive binary cross-entropy, and focal loss. Trained end-to-end with AdamW and evaluated across multiple independent domains, the model demonstrates strong generalization under scanner and staining shifts, achieving balanced accuracy of 85.0\%, ROC-AUC of 0.927, sensitivity of 89.2\%, and specificity of 80.9\% on the official test set. These results indicate that combining DenseNet-121 with stain-aware augmentation and imbalance-adaptive objectives yields a robust, domain-generalizable framework for atypical mitosis classification, supporting its potential for reliable deployment in real-world computational pathology workflows.
\end{abstract}

\vspace{-0.2cm}
\section{Introduction}
\vspace{-0.2cm}
Characterizing mitotic figures in histopathology images is a critical step in cancer grading, as atypical mitoses often indicate aggressive tumor behavior \cite{veta2015assessment,ciresan2013mitosis}. While deep learning methods have shown promise in automating mitosis recognition, their generalization is significantly hindered by \textbf{domain shift} variations in staining protocols, acquisition scanners, tissue preparation, and tumor types across laboratories \cite{tellez2019quantifying,campanella2019clinical}. Domain shift due to scanner variability and tumor heterogeneity is a key challenge in mitosis detection \cite{aubreville_comprehensive_2023}. As a result, models trained on a specific dataset frequently exhibit performance degradation when applied to unseen domains. Recent work has benchmarked deep learning and vision foundation models for atypical mitosis classification \cite{banerjee2025benchmarkingdeeplearningvision}, highlighting the importance of robust cross-domain evaluation. Cross-species datasets such as the canine mammary carcinoma WSI collection \cite{aubreville_completely_2020} have been proposed to enrich training and support generalization studies.

The \textbf{Mitosis Domain Generalization (MIDOG) 2025 challenge} \cite{ammeling_2025_15077361}, particularly Track 2, focuses on the binary classification of mitotic figures into normal and atypical categories under multi-domain variability \cite{aubreville2020mitosis}. This task is inherently challenging due to two factors: (i) the scarcity and imbalance of atypical mitotic figures relative to normal mitoses, and (ii) the large domain variations present in histopathology images \cite{johnson2019survey,zhou2022models}. To address these challenges, we propose a \textbf{DenseNet121-based framework} enhanced with two key components. First, to mitigate domain variability, we employ a \textbf{stain-aware augmentation pipeline} incorporating Macenko normalization \cite{macenko2009method} and geometric perturbations, along with a \textbf{60\% random cropping strategy} to further improve spatial generalization. Second, to tackle label imbalance, we design a \textbf{combined loss formulation} that unifies weighted binary cross-entropy and focal loss \cite{lin2017focal}, enabling the model to emphasize minority class learning while stabilizing optimization.

Our contributions are summarized as follows:
\begin{itemize}
    \item We present a robust \textbf{stain-aware and spatially augmented preprocessing pipeline} including 60\% crop-based patch perturbation to improve domain robustness.
    \item We propose an \textbf{integrated loss formulation} combining class-weighted binary cross-entropy and focal loss, enabling effective learning under strong class imbalance.
   \item We demonstrate that our \textbf{DenseNet-121} framework generalizes strongly across domains, achieving \textbf{BAcc 85.0\%}, \textbf{ROC--AUC 0.927}, \textbf{Sensitivity 89.2\%}, and \textbf{Specificity 80.9\%} on the official MIDOG25 test set.
\end{itemize}

\vspace{-0.4cm}
\section{Method}
\vspace{-0.2cm}
\subsection{Architecture Overview}

Our framework employs a DenseNet121 backbone \cite{huang2017densenet} with a single-node classification head, chosen for its efficient feature reuse and compact parameterization. The model is trained end-to-end using stain-normalized input patches and integrates a hybrid loss combining class-weighted binary cross-entropy and focal loss \cite{lin2017focal}. This design emphasizes robust representation learning under class imbalance while maintaining generalization across staining and scanner domains \cite{aubreville_comprehensive_2023}.

\vspace{-0.2cm}
\begin{figure*}[h]
\centering
\includegraphics[width=1\linewidth]{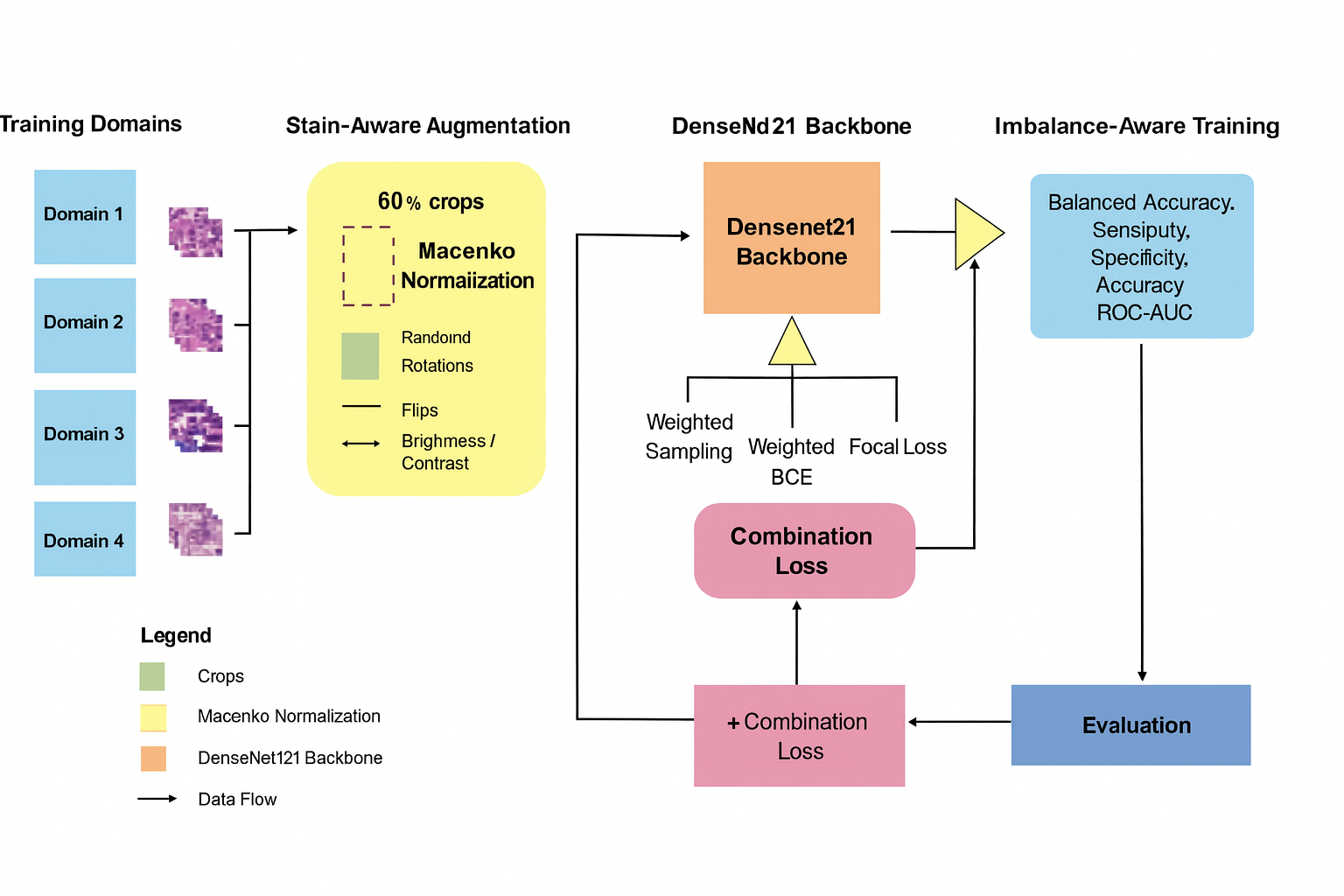}
\caption{Overview of the proposed DenseNet121-based framework for atypical mitosis classification under domain variability. The pipeline takes mitotic figure patches from multiple domains, applies stain-aware preprocessing (Macenko normalization), and uses a DenseNet121 backbone followed by a binary classification head. The training strategy integrates 60\% spatial cropping, stain perturbations, and a composite loss combining weighted cross-entropy, focal loss, and sampling-based imbalance handling. The model is evaluated using balanced accuracy, sensitivity, specificity, and ROC-AUC.}
\label{fig:method}
\end{figure*}

\subsection{Dataset}
We utilized the \textbf{MIDOG25 Atypical Mitosis Classification Training Set} \cite{weiss_2025_15188326}, derived from the AMi-Br histologic dataset \cite{palm_histologic_2025}, which provides expertly annotated mitotic figure patches categorized as atypical (AMF) or normal (NMF). This dataset spans multiple domains, each reflecting unique combinations of scanner models, staining variations, and tissue preparation protocols, thereby introducing substantial domain shift challenges \cite{aubreville_comprehensive_2023}. For internal model selection, we used an 80/20 patch-level preliminary validation split with class stratification. Because patches from the same WSI or lab/scanner may share distributional factors, such splits can leak domain cues; therefore, we anchor our claims on the official \textsc{MIDOG-25} test set and plan grouped (WSI-level) or leave-one-domain-out validation in future work.

\subsection{Preprocessing and Augmentation}
All mitotic patches were resized to $224 \times 224$ pixels. To reduce sensitivity to staining differences, we applied \textbf{stain-aware augmentation} using Macenko normalization from the TIA Toolbox. Spatial robustness was improved by introducing a \textbf{60\% random cropping} strategy during training, encouraging the model to learn localized discriminative features. Additional augmentations included random $90^{\circ}$ rotations, horizontal and vertical flips, and brightness-contrast perturbations. Validation and test samples were only resized and normalized using ImageNet statistics.

\subsection{Network Architecture}
The classification backbone is a \textbf{DenseNet121} model initialized with ImageNet-pretrained weights. The final classification layer was replaced with a single-node fully connected head for binary classification. A sigmoid activation was used during inference to convert logits into probabilities. DenseNet121 was chosen for its efficient feature reuse, compact parameterization, and strong performance on medical imaging \cite{huang2017densenet} benchmarks.

\subsection{Imbalance-Aware Optimization}
To address the scarcity of atypical mitoses and the resulting class imbalance, we adopt a hybrid objective that unifies class-weighted binary cross-entropy (WBCE) with focal loss, and we construct batches via inverse class-frequency sampling to ensure balanced exposure during training.

\paragraph{Notation.}
For sample $i$ with label $y_i\in\{0,1\}$ and model logit $z_i$, let $p_i=\sigma(z_i)$ denote the predicted probability for the positive (atypical) class. Let $w_1$ and $w_0$ be class weights for positive and negative classes, $\alpha\in[0,1]$ and $\gamma\ge 0$ the focal-loss parameters, and $\lambda\in[0,1]$ the mixing weight between losses. Mini-batch size is $B$.

\paragraph{Weighted BCE (WBCE).}
\begin{equation}
\mathrm{WBCE}(y_i,p_i) \;=\; -\Big[w_1\,y_i \log p_i \;+\; w_0\,(1-y_i)\log(1-p_i)\Big].
\label{eq:wbce}
\end{equation}

\paragraph{Focal loss.}
\begin{equation}
\mathrm{Focal}(y_i,p_i;\alpha,\gamma) \;=\; -\Big[\alpha\,y_i\,(1-p_i)^{\gamma}\log p_i \;+\; (1-\alpha)\,(1-y_i)\,p_i^{\gamma}\log(1-p_i)\Big].
\label{eq:focal}
\end{equation}

\paragraph{Combined objective.}
\begin{equation}
\mathcal{L} \;=\; \frac{1}{B}\sum_{i=1}^{B}
\Big[
\lambda\,\mathrm{WBCE}(y_i,p_i)
\;+\;
(1-\lambda)\,\mathrm{Focal}(y_i,p_i;\alpha,\gamma)
\Big].
\label{eq:combined}
\end{equation}

\paragraph{Implementation notes.}
All losses are computed from logits via a numerically stable BCE-with-logits implementation; $\lambda$ linearly mixes the two terms. We use inverse class-frequency sampling when forming mini-batches to further reduce majority-class bias.

\paragraph{Symbols and meanings.}
\begin{description}
  \item[$y_i$] Binary ground-truth label for sample $i$ ($1$ atypical
  , $0$ normal).
  \item[$z_i$] Model logit for sample $i$; $p_i=\sigma(z_i)$ is the predicted probability.
  \item[$p_i$] Predicted probability of the positive (atypical) class for sample $i$.
  \item[$w_1,w_0$] Class weights (positive/negative) used in WBCE to upweight minority-class errors.
  \item[$\alpha$] Class-balancing factor in focal loss.
  \item[$\gamma$] Focusing parameter that down-weights easy examples; larger $\gamma$ emphasizes hard samples.
  \item[$\lambda$] Mixing weight between WBCE and focal loss in the hybrid objective.
  \item[$B$] Mini-batch size; the loss is averaged over $B$ examples.
\end{description}

\subsection{Training Protocol}
Models were trained for 100 epochs using the \textbf{AdamW} optimizer  \cite{loshchilov2019adamw}  with a base learning rate of $1 \times 10^{-3}$. To encourage stable convergence, the learning rate for the DenseNet backbone was reduced by a factor of 10 compared to the classifier head. A weight decay of 0.05 was used for regularization. Batch size was set to 32, and Early stopping was applied with a patience of 50 epochs based on validation balanced accuracy.
\begin{table*}[h]
\centering
\caption{Baseline performance on preliminary test set using DenseNet121 without stain augmentation or imbalance-aware training. Balanced accuracy (BAcc) is the primary evaluation metric.}
\label{tab:baseline_results}
\begin{tabular}{llcccc}
\hline
Domain &  BAcc &Accuracy & Sensitivity & Specificity & ROC-AUC \\
\hline
0 &  0.719 &0.694 & 0.750 & 0.688 & 0.219 \\
1 &  0.822 &0.708 & 1.000 & 0.644 & 0.118 \\
2 &  0.773 &0.688 & 0.972 & 0.573 & 0.093 \\
3 &  0.903 &0.816 & 1.000 & 0.806 & 0.028 \\
\hline
\textbf{Overall} &  \textbf{0.809} &\textbf{0.711} & \textbf{0.972} & \textbf{0.647} & \textbf{0.100} \\
\hline
\end{tabular}
\end{table*}
\subsection{Evaluation Metrics}
We report the following domain-aware classification metrics:
\begin{itemize}
    \item \textbf{Balanced Accuracy (BAcc)}: Averaged recall for AMF and NMF classes; primary metric.
    \item \textbf{Sensitivity (AMF recall)}: Measures how well atypical mitoses are detected.
    \item \textbf{Specificity (NMF recall)}: Captures false positive rate on normal mitoses.
    \item \textbf{Overall Accuracy}: General correctness across all patches.
    \item \textbf{ROC-AUC}: Threshold-independent evaluation of discriminative performance.
\end{itemize}

\vspace{-0.2cm}
\section{Results}
\vspace{-0.5cm}
\begin{table*}[h]
\centering
\caption{Performance on preliminary test set of our improved DenseNet121 framework with stain-aware augmentation, 60\% cropping, and hybrid loss.}
\label{tab:improved_results}
\begin{tabular}{llcccc}
\hline
Domain &  BAcc &Accuracy & Sensitivity & Specificity & ROC-AUC \\
\hline
0 &  0.625 &0.722 & 0.500 & 0.750 & 0.695 \\
1 &  0.797 &0.733 & 0.897 & 0.697 & 0.853 \\
2 &  0.865 &0.808 & 1.000 & 0.730 & 0.936 \\
3 &  0.889 &0.789 & 1.000 & 0.778 & 0.944 \\
\hline
\textbf{Overall} &  \textbf{0.826} &\textbf{0.764} & \textbf{0.930} & \textbf{0.723} & \textbf{0.890} \\
\hline
\end{tabular}
\end{table*}

To evaluate the effectiveness of our proposed enhancements, including stain-aware augmentation, 60\% random cropping, and a hybrid loss function, we compared the performance of our baseline DenseNet121 model with the improved variant on the preliminary test set of MIDOG25 atypical mitosis classification dataset.

\subsection{Baseline Performance (Single-Head DenseNet121)}
We first evaluated a baseline configuration using DenseNet121 with a simple single-head classifier and basic cross-entropy loss, without augmentation or loss reweighting. Results across four domains are reported in Table~\ref{tab:baseline_results}.
This baseline model achieved a reasonable, balanced accuracy of \textbf{0.809}, with very high sensitivity (\textbf{0.972}), but relatively poor specificity (0.647) and extremely low ROC-AUC values across domains. This confirms that although the model could detect atypical mitoses well, it struggled with general discrimination and domain adaptation.
\subsection{Improved Performance with Full Method}
After incorporating our full pipeline stain-aware augmentation, 60\% cropping, and a combined WBCE + focal loss, the DenseNet121 model demonstrated significantly improved generalization and calibration across all domains, as shown in Table~\ref{tab:improved_results}.

Compared to the baseline, the improved model:
\begin{itemize}
    \item Increased the overall balanced accuracy from \textbf{0.809} to \textbf{0.826}.
    \item Achieved a significant gain in overall ROC-AUC (\textbf{0.890} vs. \textbf{0.100}).
    \item Maintained strong sensitivity (\textbf{0.930}), while improving specificity (\textbf{0.723}).
\end{itemize}
\begin{table*}[h]
\centering
\caption{Final performance on the MIDOG25 test set using our DenseNet121 framework with stain-aware augmentation and hybrid loss.}
\label{tab:final_results}
\begin{tabular}{lccccc}
\hline
Domain & BAcc & Accuracy & Sensitivity & Specificity & ROC-AUC \\
\hline
0  & 0.828 & 0.770 & 0.917 & 0.740 & 0.902 \\
1  & 0.967 & 0.946 & 1.000 & 0.933 & 0.995 \\
2  & 0.814 & 0.758 & 0.966 & 0.661 & 0.936 \\
3  & 0.820 & 0.822 & 0.818 & 0.822 & 0.900 \\
4  & 0.756 & 0.925 & 0.571 & 0.940 & 0.874 \\
5  & 0.838 & 0.784 & 0.944 & 0.732 & 0.934 \\
6  & 0.870 & 0.839 & 0.913 & 0.827 & 0.933 \\
7  & 0.816 & 0.814 & 0.820 & 0.812 & 0.903 \\
8  & 0.861 & 0.840 & 0.902 & 0.821 & 0.943 \\
9  & 0.789 & 0.890 & 0.677 & 0.901 & 0.887 \\
10 & 0.747 & 0.758 & 0.861 & 0.633 & 0.886 \\
11 & 0.783 & 0.715 & 0.884 & 0.682 & 0.867 \\
\hline
\textbf{Overall} & \textbf{0.850} & \textbf{0.823} & \textbf{0.892} & \textbf{0.809} & \textbf{0.927} \\
\hline
\end{tabular}
\end{table*}


\subsection{Final Test Performance}
On the official MIDOG25 test set, our method achieved an overall balanced accuracy of \textbf{0.850}, ROC-AUC of \textbf{0.927}, sensitivity of \textbf{0.892}, and specificity of \textbf{0.809}. 
Compared to the preliminary evaluation, the final results confirm that our DenseNet121 framework generalizes well across unseen domains. 
Notably, Domain~1 achieved nearly perfect performance (BAcc 0.967, ROC-AUC 0.995), while performance in challenging domains such as Domain~4 (BAcc 0.756) and Domain~10 (BAcc 0.747) highlights areas where further domain-adaptive strategies may be beneficial.


These results validate the effectiveness of our proposed enhancements, showing robust performance across all domains, improved discrimination capability, and a favorable trade-off between sensitivity and specificity. The use of a hybrid loss and stain-aware augmentation allowed the model to better generalize to unseen domain shifts, addressing key challenges in computational pathology.

\section{Discussion}
This study introduces a DenseNet121-based framework for atypical mitosis classification under domain shift, as posed by the \textsc{MIDOG25} challenge. On the official \textsc{MIDOG25} test set, the model attains a \textbf{balanced accuracy} of \textbf{0.850}, \textbf{ROC--AUC} of \textbf{0.927}, \textbf{sensitivity} of \textbf{0.892}, and \textbf{specificity} of \textbf{0.809}, demonstrating reliable generalization across unseen scanner and staining conditions.

On an internal preliminary split, the approach achieves \textbf{balanced accuracy} \textbf{0.83}, \textbf{ROC--AUC} \textbf{0.89}, and high \textbf{sensitivity} \textbf{0.93}. Relative to a baseline trained with standard procedures and \emph{without} stain-aware augmentation, the improved framework increases \textbf{specificity} (\textbf{0.72} vs.\ 0.65) and \textbf{ROC--AUC} (\textbf{0.89} vs.\ 0.10), indicating better discrimination and calibration in heterogeneous domains. We attribute these gains to three design choices: (i) stain-aware augmentation via Macenko normalization, (ii) a 60\% random crop to encourage morphological focus, and (iii) a hybrid objective combining class-weighted binary cross-entropy with focal loss.

Despite these improvements, specificity remains lower than sensitivity, reflecting a tendency toward over-detection. While this bias reduces false negatives—a desirable property for tumor grading—further reducing false positives is important for downstream efficiency. Promising directions include self-supervised pretraining on histopathology corpora, domain-adaptive or test-time adaptation techniques, and attention mechanisms to enhance specificity without sacrificing sensitivity.

Overall, coupling an efficient DenseNet121 backbone with domain-adaptive augmentation and imbalance-sensitive learning provides a strong foundation for atypical mitosis recognition and supports practical deployment in computational pathology workflows.

\section{Conclusion}
We presented a DenseNet121-based framework for atypical mitosis classification in the \textsc{MIDOG25} setting, explicitly addressing domain shift and class imbalance. By combining stain-aware augmentation (Macenko normalization), a 60\% random-cropping strategy to encourage morphological focus, and a hybrid objective unifying class-weighted binary cross-entropy with focal loss, the method achieves strong cross-domain generalization. On the \emph{official} \textsc{MIDOG25} test set, our framework attains \textbf{balanced accuracy} \textbf{0.850}, \textbf{ROC--AUC} \textbf{0.927}, \textbf{sensitivity} \textbf{0.892}, and \textbf{specificity} \textbf{0.809}. On an internal preliminary split, it reaches \textbf{balanced accuracy} \textbf{0.826}, \textbf{ROC--AUC} \textbf{0.890}, and \textbf{sensitivity} \textbf{0.930}, with improved \textbf{specificity} (\textbf{0.723}) relative to a baseline trained without stain-aware or imbalance-aware components. These findings suggest that pairing a lightweight, well-regularized backbone with domain- and imbalance-sensitive training is an effective recipe for robust atypical mitosis recognition.

\noindent\textbf{Limitations and future work.}
Although sensitivity is high, further improving specificity remains a priority to reduce false positives in downstream clinical workflows. Promising directions include: (i) self-supervised or foundation-model pretraining tailored to H\&E variability; (ii) stain- and scanner-invariant representation learning; (iii) calibration and threshold optimization for cost-sensitive deployment; (iv) explicit domain adaptation or test-time adaptation; (v) uncertainty estimation and interpretable explanations to increase clinical trust; and (vi) broader multi-institutional validation. We anticipate these extensions will further enhance reliability and facilitate real-world adoption.

\bibliographystyle{splncs04}
\bibliography{main}
\end{document}